\documentclass[letterpaper, 10 pt, conference]{ieeeconf}  % Comment this line out if you need a4paper

\IEEEoverridecommandlockouts                              % This command is only needed if 
\usepackage{amsmath} % assumes amsmath package installed
\usepackage{amssymb}  % assumes amsmath package installed
\usepackage{algorithm}
\usepackage{algpseudocode}
\usepackage{tabularx}
\usepackage{graphicx}
\usepackage{mathtools}
\usepackage{gensymb}
\usepackage{xcolor}

\newcommand{\revise}[1]{\textcolor{black}{#1}}

\title{\LARGE \bf
Deployable Vision-driven UAV River Navigation via Human-in-the-loop Preference Alignment
}

\author{Zihan Wang$^{1}$, Jianwen Li$^{1}$, Li-Fan Wu$^{1}$ and Nina Mahmoudian$^{1}$
\thanks{This work was supported by N00014-24-1-2019.}% <-this % stops a space
\thanks{$^{1}$Zihan Wang, Jianwen Li, Li-Fan Wu and Nina Mahmoudian are with the School of Mechanical Engineering, Purdue University, West Lafayette, IN 47907, USA
        {\tt\small wang5044, li3602, wu1714, ninam@purdue.edu}}%
}

\begin{document}

\maketitle
\thispagestyle{empty}
\pagestyle{empty}

%%%%%%%%%%%%%%%%%%%%%%%%%%%%%%%%%%%%%%%%%%%%%%%%%%%%%%%%%%%%%%%%%%%%%%%%%%%%%%%%
\begin{abstract}
Rivers are critical corridors for environmental monitoring and disaster response, where Unmanned Aerial Vehicles (UAVs) guided by vision-driven policies can provide fast, low-cost coverage.
However, deployment exposes simulation-trained policies with distribution shift and safety risks and requires efficient adaptation from limited human interventions.
We study human-in-the-loop (HITL) learning with a conservative overseer who vetoes unsafe or inefficient actions and provides statewise preferences by comparing the agent’s proposal with a corrective override. 
We introduce Statewise Hybrid Preference Alignment for Robotics (SPAR-H), which fuses direct preference optimization on policy logits with a reward-based pathway that trains an immediate-reward estimator from the same preferences and updates the policy using a trust-region surrogate. 
With five HITL rollouts collected from a fixed novice policy, SPAR-H achieves the highest final episodic reward and the lowest variance across initial conditions among tested methods.
The learned reward model aligns with human-preferred actions and elevates nearby non-intervened choices, supporting stable propagation of improvements. 
We benchmark SPAR-H against imitation learning (IL), direct preference variants, and evaluative reinforcement learning (RL) in the HITL setting, and demonstrate real-world feasibility of continual preference alignment for UAV river following. 
Overall, dual statewise preferences empirically provide a practical route to data-efficient online adaptation in riverine navigation.

\end{abstract}

%%%%%%%%%%%%%%%%%%%%%%%%%%%%%%%%%%%%%%%%%%%%%%%%%%%%%%%%%%%%%%%%%%%%%%%%%%%%%%%%
\section{INTRODUCTION}
% Motivate drone river following
Rivers are natural corridors for aerial robots used in environmental monitoring, search and rescue, hydrological surveying, and infrastructure inspection. 
An Unmanned Aerial Vehicle (UAV) that can reliably follow the river unlocks long-range coverage with minimal prior maps. 
Vision-driven navigation is especially attractive in riverine settings where GNSS can degrade near bridges and canopies. 
Recent work shows that first-person–view policies can be learned to react to water cues, using semantic segmentation or end-to-end reactive control, often with simulation pretraining \cite{ross2013learning, giusti2015machine, loquercio2018dronet, wei2022vision, wang2025vision}. 

% Motivate HITL
However, riverine scenes are dynamic (bridges, tributaries, specular water), which makes fixed, hand-tuned rewards in simulation brittle and costly to maintain when deployed to field \cite{marta2021human}.
Policies pretrained in simulation also face distribution shift in the field, creating performance and safety risks \cite{kaufmann2024survey, korkmaz2025mile}.
These factors motivate a human-in-the-loop (HITL) regime in which an overseer vets proposed actions, vetoes unsafe or inefficient choices, and provides sparse corrections \cite{xu2022look, choi2020fast}.
Because operator time is limited, the central challenge is to learn efficiently from this feedback during deployment, turning a small number of interventions into reliable improvements.

\begin{figure}[t]
    \centering
    \vspace{0.16cm}
    \includegraphics[width=\linewidth]{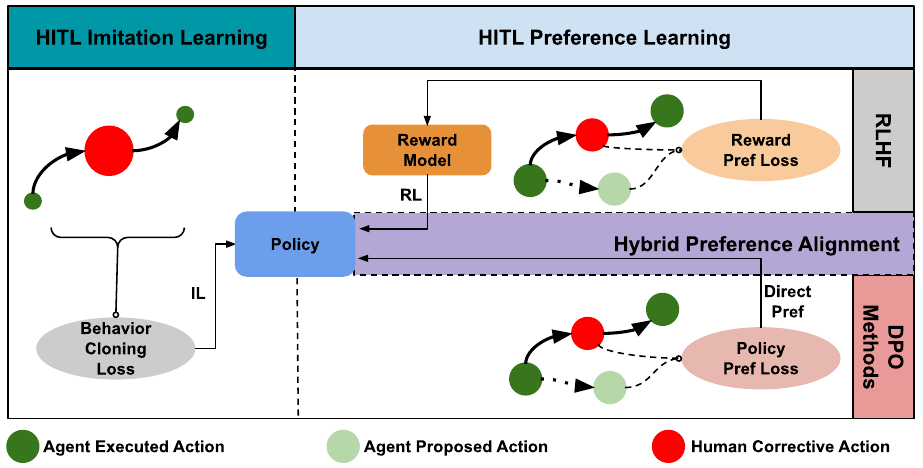}
    \caption{Comparison of HITL learning methods. Statewise hybrid preference alignment combines direct (RL-free) preference optimization applied to policy logits and RLHF-style reward model preferences that indirectly drive policy update. Both signals arise from per-state comparisons between the human override and the agent proposal. Imitation learning methods mainly use weighted behavior cloning on the human-intervened trajectory, where human corrective actions are given larger weights. 
    % Best view in color.
    }
    \label{fig:spar-h-diagram}
\end{figure}
% \vspace{-0.1 in}

% Reason to use HITL
% Rewards (and optionally costs in safe RL) are the major sources of signals that drive the learning of a RL agent. However, specifying the reward function can be challenging or impractical \cite{marta2021human}, especially in vision-driven tasks where the observation space of the agent is of high dimension, task objectives are not directly correlated with visual inputs \cite{wang2024synergistic, wang2024vision}, or in real environments where numerical reward feedback is not easily accessible.
% Moreover, distributional shifts from simulation to real world, as well as over-optimization in simulation, might degrade RL policy performance in real world deployments, causing practical and safety concerns \cite{kaufmann2024survey, korkmaz2025mile}.
% To circumvent challenging reward engineering and foster adaptability of RL policy in real-world dynamic environments, it is inevitable to incorporate interactive human interventions in the deployment \cite{xu2022look}, while letting the agent to learn from the limited human feedback \cite{choi2020fast}, in the human-in-the-loop (HITL) learning fashion.

% HITL families and limitations
% HITL learning in robotics typically spans three families. 
We organize HITL learning in robotics by how feedback is consumed, Fig.~\ref{fig:spar-h-diagram}.
\emph{Imitation learning (IL)} clones corrective actions on intervened segments (e.g., IWR \cite{mandlekar2020human}, HG-DAgger \cite{kelly2019hg}, Sirius \cite{liu2022robot}, and \cite{ou2023towards}), but does not exploit the same-state comparison between the agent's proposal and the human's correction. 
\emph{Reinforcement Learning from Human Feedback (RLHF)} first learns a scalar reward or value from human feedback and then updates the policy via RL.
Examples include rewards inferred from trajectory comparisons or demonstrations \cite{christiano2017deep, ibarz2018reward}, online and query-efficient preference collection \cite{lee2021pebble}, and signals derived from interventions or safety events such as intervention-shaped values \cite{peng2022safe} and rewards assigned from intervention flags \cite{luo2023rlif}. 
Most of these focus on trajectory- or episode-level preferences rather than specific intervention states, facing temporal credit assignment issue, especially under partial observability \cite{wirth2017survey}.
\emph{Direct preference optimization (DPO, \cite{rafailov2023direct})}, or \emph{Bradley--Terry (BT, \cite{bradley1952rank})} preference model, can be used to apply a preference loss directly to the policy logits at the intervened states (e.g., \cite{mengpoil, xia2025robotic, khorasani2025fusing}) without a reward model, but typically requires extra regularization terms and offers limited propagation to nearby, non-intervened states.

% State the gap
\revise{Moreover, much of this literature is evaluated in short-horizon, frequently reset manipulation or locomotion settings. 
In contrast, vision-driven river following is long-horizon and first-person partially observable, where local images do not reveal global progress, and preprocessing can misinterpret water, bridges, or reflections, introducing nonstationary perception noise. 
In addition, preference supervision is typically trajectory- or segment-level, and many systems rely on biased sampling from replay buffers, making it difficult to disentangle algorithmic effects from sampling effects. 
Statewise preferences captured exactly at human intervention points under long-range partial observability remain comparatively under-explored in these conditions.}

In light of these limitations, we propose the statewise hybrid preference alignment for robotics (SPAR-H): for each intervention, the per-state comparison between the agent-proposed action and the human-corrected action drives two pathways—a direct BT loss on policy logits (RL-free) and an indirect BT loss on an immediate-reward estimator whose advantages update the policy with a trust-region RL surrogate on non-intervened steps (RLHF-style), Fig.~\ref{fig:spar-h-diagram}. 
This dual signal preserves the strong in-distribution corrections of direct preference learning while propagating improvements beyond the intervened states.

Our study evaluates these HITL learning methods \revise{in their simplest forms} in vision-driven river following task under a controlled \revise{(full replay buffer is used)}, budgeted HITL protocol.
% , complementing prior evaluations that emphasize gridworld, manipulation, or locomotion tasks.
In simulation, SPAR-H attains the highest episodic reward and the lowest standard deviation at the final retrained checkpoint. 
The reward estimator improves alignment with human preferences over training. 
\revise{
We demonstrate the first real-world UAV deployment of statewise hybrid preference learning, showing SPAR-H adapts policies online during river following despite perception imperfections.
}
% We further demonstrate real-world feasibility on a river-following drone, indicating that SPAR-H can adapt policies online under perception imperfections during deployment.
\revise{
Our work extends preference alignment techniques to this setting, empirically showing that SPAR-H can enable both local correction and broader policy adaptation under limited feedback budgets.
}
% Main contributions
To this end, our contributions are:
\begin{itemize}
    \item A unified HITL framework that turns statewise human corrections into both direct policy updates and reward-based RL targets in a single model.
    \item A controlled evaluation on vision-driven river following with human interventions, comparing direct statewise preference, RLHF, IL, and evaluative RL methods under a fixed feedback budget.
    \item A real-world deployment of our HITL preference learning stack on a UAV river following task, demonstrating rapid online adaptation from sparse corrections and declining interventions under imperfect perception.
\end{itemize}

\section{PRELIMINARIES}

% MDP
\subsection{Markov Decision Process}\label{sec:prelim_mdp}
We focus on Markov Decision Process (MDP) with discrete state and action spaces, defined as $(S, A, R, T, \mu, \gamma)$, where $S$ is the set of states, $A$ is the set of actions, $R: S \times A \rightarrow \mathbb{R}$ is the reward function, $T: S \times A \rightarrow S$ is the state transition function, and $\mu: S \rightarrow [0, 1]$ is the initial state distribution, $\gamma \rightarrow [0, 1]$ is the discount factor.
Starting from an initial state $s_0 \sim \mu$, at each time stamp, the agent chooses an action $a_t \in A$ according to the current policy $\pi_{\theta} (s_t)$ parameterized by $\theta$, transitions to the next state $s_{t+1} = T(s_t, a_t)$, and receives a reward $r_t = R(s_t, a_t)$. 
The objective of RL is to learn an optimal policy $\pi^*_{\theta}$ that maximizes the expected cumulative discounted rewards 
\begin{equation}\label{eqn:rl-objective}
    \theta^* = \arg \max_{\theta} \mathbb{E}_{s_0 \sim \mu} [\sum_{t=0}^{\infty} \gamma^t R(s_t, a_t)].
\end{equation}
%In the vision-driven drone navigation application, the state $s$ is the first-person-view image, which renders the setting to be partially observable MDP (POMDP) due to the independence on positioning sensors of our approach. 

% Vision processing + framework
\begin{figure*}[ht]
    \centering
    \vspace{0.15cm}\includegraphics[width=0.98\textwidth]{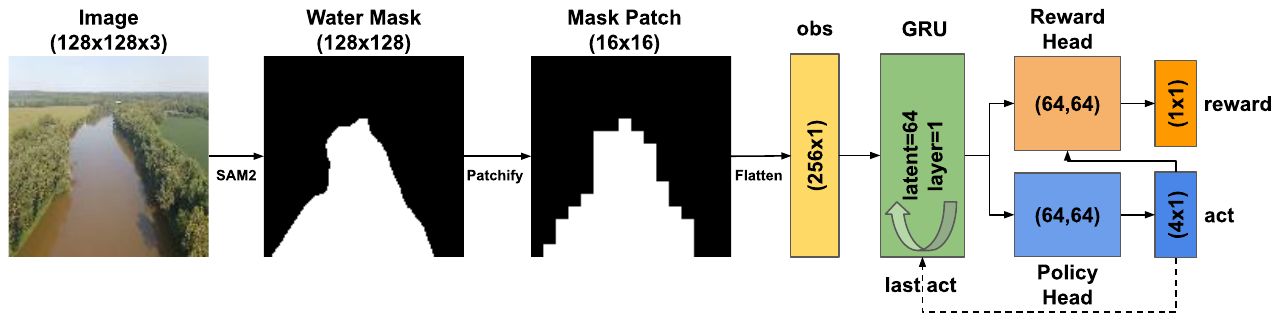}
    \caption{Vision-to-action pipeline for river following. RGB is segmented by SAM2 into a water mask, patchified, passed through a frozen GRU encoder, then split to a policy head (action) and a reward head (immediate reward).}
    \label{fig:arch}
\end{figure*}

% Submodular POMDP
\subsection{Submodular Rewards under Partial Observability}\label{sec:prelim_spomdp}

% River navigation, as a coverage control problem, exhibits diminishing returns: the incremental value of revisiting already covered regions decreases as coverage grows. 
% This is captured by a monotone submodular set function $F:2^{\mathcal{S}}\!\to\!\mathbb{R}$ over visited state--time pairs, where the marginal gain $\Delta(x\mid S_1)=F(A\cup\{x\})-F(A)$ decreases as the visited set $A$ expands. 
% In a submodular MDP \cite{prajapat2023submodular}, the episode reward is $F(\tau)$, i.e., the value of the set of states visited by trajectory $\tau$; this makes reward \emph{history-dependent} (non-additive) even when dynamics are Markovian, and includes coverage and informative path planning as special cases.

% What are submodularity and submodular reward
\textbf{Submodular Rewards.}
River navigation is a coverage-control problem with diminishing returns: agent will be rewarded only if it visits un-visited river segments \cite{wang2024synergistic, wang2024vision, wang2025vision}.
Formally, we discretize the river into elements $\mathcal{V}$ and let $C_t \subseteq \mathcal{V}$ be the set of visited elements up to time $t$. 
Let $F:2^{\mathcal{V}}\!\to\!\mathbb{R}$ be a monotone submodular coverage utility \cite{prajapat2023submodular}, with marginal gain
$\Delta_F(v \mid C) \;=\; F(C\cup\{v\}) - F(C)$.
For any $C_1 \subseteq C_2 \subseteq \mathcal{V}$ and any $v \in \mathcal{V}\setminus C_2$, 
$\Delta_F(v \mid C_1) \;\ge\; \Delta_F(v \mid C_2).$
% \begin{equation}\label{eqn:subm}
%     \Delta_F(v \mid C_1) \;\ge\; \Delta_F(v \mid C_2). 
% \end{equation}
The episode reward of an episode with length $T$ is $F(C_T)$ and the per\mbox{-}step submodular reward is the marginal gain
$r_t = \Delta_F(v_t \mid C_{t-1})$ that the immediate-reward estimator $R_{\phi}$ tries to converge to (more in Section \ref{sec:approach_spar-h}).
In other words, revisits yield zero gain while entering a new element yields positive gain.

% How submodularity and partial observability affect the design of the learning framework?
\textbf{Partial Observability.} 
A first-person image lacks global coverage context (which river segments have already been visited), which renders the setting to be partially observable MDP (POMDP).
However, both the optimal policy and the immediate reward depend on this visitation history rather than the current image alone.
For example, two nearly identical observations can yield different rewards depending on whether the latest motion advanced coverage or stalled. 
We therefore maintain a recurrent latent state
% $z_t = g_\psi\big(z_{t-1},\,o_t,\,a_{t-1}\big)$
$z_t = g\big(z_{t-1},\,o_t,\,a_{t-1}\big)$
via a Gated Recurrent Unit (GRU \cite{chung2014empirical}) and learn an immediate-reward estimator $R_\phi(z_t,a_t)$ that approximates the marginal gain $\Delta(\cdot \mid z)$\protect\footnotemark[1], shown in Fig.~\ref{fig:arch}. 
Here $z$ is the learned representation of the set of visited segments $C$.

\footnotetext[1]{
In our setup $s_t=z_t$ (the GRU latent). Hereafter, we use $s_t$ to stay model-agnostic: in fully observable or feed-forward cases $s_t=o_t$, or any learned context conditioning policy or reward head.
}

% Notation explanation to remove ambiguity
% \textbf{Notation Convention.}
% We use $s_t$ to denote the input consumed by both the policy and the reward heads in the following sections. 
% In our implementation this input is the GRU latent, so $s_t = z_t$. 
% We choose $s_t$ over $z_t$ to keep the notation standard and to emphasize that the method is agnostic to the encoder. 
% In fully observable or feed-forward variants one may take $s_t = o_t$, and in other architectures $s_t$ can be any (learned) contextual representation that conditions both $\pi_\theta$ and $R_\phi$.

%This architecture choice—one GRU encoder with a policy head and an immediate-reward head—explains why we do not use a separate value critic with GAE in later sections. In submodular, history-dependent problems a value function on raw observations $V(o_t)$ is ill-posed; even a recurrent critic must implicitly reconstruct the visited set. By predicting immediate marginal gains with $R_\phi$ and summarizing history in $z_t$, we obtain advantages that are faithful to the submodular objective while keeping the learning problem well-posed under partial observability.

% BT preference model
\subsection{Preference Model}\label{sec:prelim_bt}
Given a context $s$ and two candidate outputs $y^+, y^- \in \mathcal{Y}(s)$ (for example, actions at state $s$ or short trajectory segments), let $U_\psi(s,y)\in\mathbb{R}$ be any scalar scoring function that assigns a utility to candidate $y$ under context $s$. We use the canonical Bradley--Terry (BT, \cite{bradley1952rank}) model to specify the probability that $y^+$ is preferred to $y^-$ as a logistic function of the score difference:
\begin{equation}
\label{eqn:bt}
p\!\left(y^+ \succ y^- \mid s\right)
= \sigma\!\big(\,\beta\,\big[U_\psi(s,y^+) - U_\psi(s,y^-)\big]\big),
\end{equation}
where $\sigma(z)=1/(1+e^{-z})$ and $\beta>0$ is an inverse-temperature.
% (often set to 1 or absorbed into $U_\psi$). 
Given a dataset $\mathcal{D}=\{(s_i, y_i^+, y_i^-)\}_{i=1}^N$ of pairwise preferences, the negative log-likelihood objective is
\begin{equation}
\label{eqn:bt_loss}
\mathcal{L}_{\mathrm{BT}}(\psi)
= - \sum_{i=1}^N \log \sigma\!\big(\,\beta\,\big[U_\psi(s_i,y_i^+) - U_\psi(s_i,y_i^-)\big]\big).
\end{equation}
This objective encourages larger scores for preferred items while decreasing scores of the dis-preferred ones. 
% In later sections we instantiate $U_\psi$ in two ways that share this same BT likelihood: (i) a learned reward estimator, $U_\psi(s,a)=R_\phi(s,a)$, which yields preference learning for reward modeling; and (ii) a policy-based score, $U_\psi(s,a)=\log \pi_\theta(a\mid s)$, which yields direct preference optimization of the policy. 
% The BT formulation is identical in both cases because it only requires a real-valued score for each candidate.

\section{APPROACH}\label{sec:approach}
% Background
\subsection{Human-in-the-loop Intervention Model}\label{sec:approach-hitl}
At time $t$, the agent observes $s_t$ and proposes $a^{\text{agent}}_t \sim \pi_\theta(\cdot \mid s_t)$. 
A human overseer either executes it or overrides with $a^{\text{human}}_t$. 
Let $m_t \in \{0,1\}$ denote an intervention. The executed action is
\begin{equation}
a^{\text{exec}}_t \;=\;
\begin{cases}
a^{\text{human}}_t & \text{if } m_t=1,\\[2pt]
a^{\text{agent}}_t & \text{if } m_t=0,
\end{cases}
\end{equation}
and we log a same-state preference with an override flag $(s_t,\,a^{\text{human}}_t \succ a^{\text{agent}}_t, m_t)$ at each step.
Hereafter, we will use $a^{\text{h}}$, $a^{\text{a}}$ and $a^{\text{e}}$ to denote $a^{\text{human}}$, $a^{\text{agent}}$ and $a^{\text{exec}}$ for brevity.

\subsection{Statewise Hybrid Preference Alignment for Robotics}\label{sec:approach_spar-h}
Human corrections at intervention steps define statewise preferences
\begin{equation}\label{eqn:pref_dataset}
    \mathcal{D}_{\mathrm{pref}}=\{(s_t, a_t^{\mathrm{h}} \succ a_t^{\mathrm{a}})\;|\; m_t=1,\; a_t^{\mathrm{h}}\neq a_t^{\mathrm{a}}\}.
\end{equation}
We use a shared recurrent encoder feeding a policy head $\pi_\theta(s)$ and an immediate-reward head $R_\phi(s,a)$, Fig.~\ref{fig:arch}. 
Statewise Preference Alignment for Robotics (SPAR) refers to how these preferences supervise the two heads. 
SPAR is a framework label, not a new loss.

\textbf{Direct path (SPAR-P)}: BT on policy logits at intervened states\revise{: $U_\psi(s,a)=\log \pi_\theta(a\mid s)$}.
% For multi-discrete actions we use the joint log-probability $\log\pi_\theta(a\mid s)=\sum_k \log\pi^{(k)}_\theta(a^{(k)}\mid s)$. 
The direct objective is the BT negative log-likelihood applied to policy logits:
\begin{equation}
\label{eqn:spar-p}
    \mathcal{L}_{\mathrm{SPAR-P}}(\theta)
    = -\!\!\sum_{\mathclap{(s,a^{\mathrm{h}}\succ a^{\mathrm{a}})\in\mathcal{D}_{\mathrm{pref}}}}
    \log \sigma\!\Big(\log\pi_\theta(a^{\mathrm{h}}\!\mid s) - \log\pi_\theta(a^{\mathrm{a}}\!\mid s)\Big).
\end{equation}

\textbf{Indirect path (SPAR-R)}: BT on immediate reward estimator \revise{: $U_\psi(s,a)=R_\phi(s,a)$}, plus trust-region RL on non-intervened steps.
The same preferences train the immediate-reward estimator with BT:
\begin{equation}
\label{eqn:spar-r}
% \mathcal{L}_{\mathrm{SPAR-R}}(\phi)
% = -\sum_{(s,a^{\text{h}}\succ a^{\text{a}})\in\mathcal{D}_{\mathrm{pref}}}
% \log \sigma\!\big(R_\phi(s,a^{\text{h}})-R_\phi(s,a^{\text{a}})\big).
\mathcal{L}_{\mathrm{SPAR\text{-}R}}(\phi)
= -\sum_{\mathclap{(s,\,a^{h}\succ a^{a})\in\mathcal{D}_{\mathrm{pref}}}}
\log \sigma\!\big(R_\phi(s,a^{h})-R_\phi(s,a^{a})\big).
\end{equation}
We update $\pi_\theta$ on non-intervened steps ($m_t{=}0$) with a trust-region surrogate \revise{to ensure stable updates under off-policy corrections}.
% We restrict the RL update to non-intervened steps because human overrides make the data off-policy, and would yield unstable importance sampling ratios and KL signals. 
% Using $m_t{=}0$ keeps the update on the agent's own action distribution, so advantages reflect what the recurrent policy actually executed.
For trust-region policy loss, we follow the work done in \cite{wang2025vision}, and adopt First Order Optimization in Policy Space (FOCOPS, \cite{zhang2020first}).
The per-state KL indicator in FOCOPS provides a hard update gate that prevents large policy moves when the reference divergence is high, yielding stable first-order updates with minimal computation.
% , and it integrates cleanly with our importance ratio and multi-discrete action factorization. 
% This makes FOCOPS a practical choice for online HITL updates under a learned immediate-reward signal.

% For a multi-discrete policy, we use the joint log-probability $\log\pi_\theta(a\mid s)=\sum_i \log\pi^{(i)}_\theta(a^{(i)}\mid s)$, where $i$ denotes certain branch of a multi-discrete action. 
For each step, define the importance ratio
\begin{equation}
\label{eqn:ratio}
\rho_t(\theta;\,s_t,a^{\mathrm{e}}_t)
=
\exp\!\Big(\log \pi_\theta(a^{\mathrm{e}}_t \mid s_t) 
            - \log \pi_{\theta_0}(a^{\mathrm{e}}_t \mid s_t)\Big),
\end{equation}
where $\pi_{\theta_0}$ is the frozen reference policy at the start of the update. 
Given an episodic trajectory $(s_t,a^{\text{e}}_t,\dots,s_{T-1},a^{\text{e}}_{T-1})$, 
we use a truncated $K$-step reward--to--go \cite{mnih2016asynchronous} \revise{(due to independence on value critic, Section \ref{sec:prelim_spomdp})} under the learned immediate reward and then standardize within the episode:
\begin{equation}
\label{eqn:rtg}
% \hat G_t \;=\; \sum_{k=0}^{T-t-1}\gamma^{k}\,R_\phi(s_{t+k},a^{\mathrm{e}}_{t+k}),
\hat G^{(K)}_t \;=\; \sum_{k=0}^{K_t-1}\gamma^{k}\,R_\phi\!\big(s_{t+k},a^{\mathrm{e}}_{t+k}\big),
\quad
A_t \;=\; \frac{\hat G_t-\bar G}{\sigma_G+\varepsilon},
\end{equation}
where $K_t := \min\{K,\,T-t\}$, $\gamma$ is the discount factor, $\bar G$ and $\sigma_G$ are the batch mean and standard deviation of $\hat G^{(K)}$, and $\varepsilon>0$ ensures numerical stability. 
% Subtracting the baseline $\bar G$ preserves the unbiased REINFORCE gradient estimator \cite{williams1992simple}. 
% Truncation to $K$ steps trades bias for reduced variance and limits error propagation from a nonstationary $R_\phi$. 
% The division by $\sigma_G$ is a standardization heuristic, which does not affect the expected gradient direction but stabilizes advantage magnitudes across batches.
% Centering by $\bar G$ is a constant baseline that preserves the unbiased policy--gradient estimator \cite{williams1992simple}, dividing by $\sigma_G$ rescales the update.
% We adopt this reward advantage estimation in place of value-based advantages because SPAR relies on an immediate--reward estimator $R_\phi$ rather than a value critic, avoiding bootstrapping under partial observability and submodular rewards (Section \ref{sec:prelim_spomdp}). 

The FOCOPS policy loss using reward-to-go advantages 
% computed from the learned immediate-reward estimator $R_\phi$ 
is
\begin{equation}
\label{eqn:focops_loss}
\mathcal{L}^{R_\phi}_{\text{FOCOPS}}(\theta) \;=\;
\mathbb{E}\!\left[ \mathbf{1}\{\mathrm{KL}_t \le \eta\}\,
\Big(\mathrm{KL}_t - \tfrac{1}{\lambda}\, \rho_t\, A_t\Big) \right],
\end{equation}
where $\mathrm{KL}_t=D_{\mathrm{KL}}\!\big(\pi_\theta(\cdot \mid s_t) \| \pi_{\theta_0}(\cdot \mid s_t)\big)$ measures the Kullback–Leibler (KL) divergence from the most recent policy to the reference policy before update, $\mathbf{1}\{\cdot\}$ is the per-state indicator function to restrain large policy gradients 
% due to first order approximation approach in FOCOPS
, $\eta$ is the threshold of KL divergence, and $\lambda$ is a hyperparameter related to the greediness of FOCOPS algorithm.

\textbf{Hybrid objective (SPAR-H, proposed).}
The hybrid variant optimizes both pathways under their natural masks:
\begin{equation}
\label{eqn:spar-h}
\mathcal{L}_{\mathrm{SPAR-H}}(\theta)
= \underbrace{\mathcal{L}_{\mathrm{SPAR-P}}(\theta)}_{\text{applied on }m_t=1}
\;+\; \alpha
\underbrace{\mathcal{L}^{R_\phi}_{\mathrm{FOCOPS}}(\theta)}_{\text{applied on }m_t=0},
\end{equation}
where $\alpha$ is a hyperparameter that balances the two losses.
Gradients from $\mathcal{L}_{\mathrm{SPAR-P}}$ update only the policy head. 
Gradients from $\mathcal{L}_{\mathrm{SPAR-R}}$ update only the reward head. 
$\mathcal{L}^{R_\phi}_{\mathrm{FOCOPS}}$ updates the policy head on non-intervened steps. 
% We normalize advantages in Eq.~\eqref{eqn:rtg} per batch to stabilize updates and snapshot $\pi_{\theta_0}$ at episode boundaries.
We snapshot $\pi_{\theta}$ at episode boundaries.
\revise{
The novelty is in combining SPAR-P and SPAR-R into a unified online HITL framework with a hybrid objective.
}
The full procedure of SPAR-H is shown in Algorithm~\ref{alg:spar-h}.
\revise{
In plain terms, SPAR-H collects human interventions during deployment, converts them into preference pairs, and then updates both the policy and a learned reward estimator. 
The policy is adjusted directly at intervention points and indirectly through the reward estimator at non-intervened states, ensuring both precise corrections and broader propagation of alignment.
}

\begin{algorithm}[h]
\caption{SPAR-H}
\label{alg:spar-h}
\begin{algorithmic}[1]
\Require Pretrained $R_\phi$ and novice policy $\pi_\theta$, discount $\gamma$, trust-region $(\eta,\lambda)$, inner retrain epochs $E$, HITL buffer $\mathcal{D}$

\State $\pi_{\theta_0} \leftarrow \mathrm{snapshot}(\pi_\theta)$
\For{episode \textbf{ep} $= 1,2,\dots$}
  \State Roll out with HITL; collect $(s_t,a^{\mathrm{a}}_t,a^{\mathrm{e}}_t,m_t)$ and append to $\mathcal{D}$
  
  \State Extract preference pairs from $\mathcal{D}$ to form  $\mathcal{D}_{\mathrm{pref}}$ based on Eq.~\eqref{eqn:pref_dataset}
  
  \For{$e=1$ \textbf{to} $E$}
    \State \textbf{Reward BT:} update $\phi$ on $\mathcal{D}_{\mathrm{pref}}$ via Eq.~\eqref{eqn:spar-r} 
    
    \State Predict $r_t \leftarrow R_\phi(s_t, a^{\mathrm{e}}_t)$ on $\mathcal{D} \setminus \mathcal{D}_{\mathrm{pref}}$ and compute $A_t$ by Eq.~\eqref{eqn:rtg}

    \State Calculate KL divergence and importance ratio (Eq.~\eqref{eqn:ratio}) between $\pi_{\theta}$ and $\pi_{\theta_{ep-1}}$

    \State \textbf{Policy RL:} construct FOCOPS-style RL loss $\mathcal{L}^{R_\phi}_{\text{FOCOPS}}(\theta)$ via Eq.~\eqref{eqn:focops_loss}

    \State \textbf{Policy BT:} construct BT loss $\mathcal{L}_{\mathrm{SPAR-P}}(\theta)$ on policy logits from $\mathcal{D}_{\mathrm{pref}}$ via Eq.~\eqref{eqn:spar-p}

    \State \textbf{SPAR-H:} minimize the hybrid loss $\mathcal{L}_{\mathrm{SPAR-H}}(\theta,\phi)$ in Eq.~\eqref{eqn:spar-h}
    
  \EndFor
  \State $\pi_{\theta_{ep}} \leftarrow \mathrm{snapshot}(\pi_\theta)$
\EndFor
\end{algorithmic}
\end{algorithm}

% Comparison with the baselines: IL-style (IWR, HG-Dagger), policy-level preference (BT, DPO), and RLHF-style. 
\subsection{Methods Compared}\label{sec:approach_methods}
% Motivate
We compare SPAR-H against methods that span the major ways human input can shape a policy: imitation learning from corrective actions, direct preference optimization from pairwise preferences without a reward model, indirect preference optimization from pairwise preferences via a reward model, and reinforcement learning from evaluative scalar feedback.

\begin{table*}[h]
\centering
\vspace{0.2cm}
\caption{Compact comparison across implemented HITL learning methods and our SPAR variants. 
% SPAR denotes our statewise preference alignment framework. 
% SPAR-P/R/D are canonical instantiations we implement within this framework for comparison, and SPAR-H is the hybrid instantiation introduced in this work.
% “Direct pref” denotes direct preference optimization without a learned reward model. 
% The “Feedback pathway” indicates whether human feedback is applied via a learned reward model followed by RL, or applied directly to the policy.
}
\label{tab:spar_comp}
\begin{tabular}{|l|c|c|c|c|}
\hline
\textbf{Method} & \textbf{Feedback type} & \textbf{Uses preference} & \textbf{Paradigm} & \textbf{Feedback pathway} \\
\hline
\textbf{SPAR-H (proposed)} & Corrective & Yes & Hybrid (Direct pref + RL) & direct to policy \textbf{and} via reward model \\
\hline
SPAR-P (BT on policy)  & Corrective & Yes & Direct pref & direct to policy \\
\hline
SPAR-R (BT on reward $\rightarrow$ RL) & Corrective & Yes & RL & via reward model \\
\hline
SPAR-D (DPO on policy) & Corrective & Yes & Direct pref & direct to policy \\
\hline
IWR \cite{mandlekar2020human} & Corrective & No & IL & direct to policy \\
\hline
HG-DAgger \cite{kelly2019hg} & Corrective & No & IL & direct to policy \\
\hline
COACH \cite{macglashan2017interactive} & Evaluative & No & RL & direct to policy \\
\hline
\end{tabular}
\end{table*}
% \vspace{-0.1 in}

\textbf{IWR} \cite{mandlekar2020human}, or intervention weighted regression, learns by behavior cloning while emphasizing intervention segments so the policy focuses on bottlenecks. 
Let $a^{\text{h}}_t$ be the corrective action when the overseer intervenes, and let $w_t$ up\mbox{-}weight those steps. 
The objective minimizes a weighted negative log\mbox{-}likelihood
\begin{equation}
\min_\theta\ \sum_t w_t\,\ell_{\mathrm{BC}}\big(\pi_\theta(\cdot\mid s_t),\,a^{\text{h}}_t\big),
\end{equation}
where $\ell_{\mathrm{BC}}$ is cross\mbox{-}entropy loss. 
In our implementation, following IWR, we upweight each takeover example by the ratio of non-intervened to intervened samples in the batch, and keep non-intervened samples at weight one.
% In our implementation the weights are realized by balanced sampling of intervention vs.\ on\mbox{-}policy segments. 
Unlike SPAR-H, IWR does not form preference pairs or learn a reward model; it directly copies the overseer’s actions.

\textbf{HG\mbox{-}DAgger} \cite{kelly2019hg}, or human\mbox{-}gated dataset aggregation, clones the overseer only when a risk gate triggers a takeover. 
If $m_t$ indicates a takeover, the learning objective reduces to
\begin{equation}
\min_\theta\ \sum_{t:\,m_t=1}\ell_{\mathrm{BC}}\big(\pi_\theta(\cdot\mid s_t),\,a^{\text{h}}_t\big).
\end{equation}
We use a human or heuristic gate rather than training a separate risk predictor to keep the baseline lean. 
Unlike SPAR-H, HG\mbox{-}DAgger remains purely imitation on gated states and does not convert takeovers into preferences or a learned reward.

% \textbf{BT \cite{bradley1952rank} on actions} is applied directly to the policy by comparing the joint log\mbox{-}probabilities of the human and agent proposals at the same state. 
% With the joint log\mbox{-}probability $\log\pi_\theta(a\mid s)=\sum_k \log\pi^{(k)}_\theta(a^{(k)}\mid s)$ for multi\mbox{-}branch actions, the objective is
% \begin{equation}
% \min_\theta\ -\sum \log\sigma\!\Big(\log\pi_\theta(a^{\text{human}}\!\mid s)-\log\pi_\theta(a^{\text{agent}}\!\mid s)\Big).
% \end{equation}
% This baseline updates the policy directly from preferences, whereas CAPER uses the same pairs only to train a reward estimator and performs RL under that reward.

\textbf{SPAR-P} and \textbf{SPAR-R} use the same statewise preference pairs but apply them differently. 
SPAR-P performs direct preference optimization by placing a BT margin on the policy logits at intervened states ($m_t{=}1$), updating $\pi_\theta$ without a reward model (Eq.~\eqref{eqn:spar-p}). 
This yields sharp, local alignment but offers limited propagation to nearby, non-intervened states and typically benefits from additional regularization. 
SPAR-R trains an immediate-reward head $R_\phi(s,a)$ with the same BT objective on the intervened pairs, then updates the policy indirectly on non-intervened steps ($m_t{=}0$) using a trust-region RL surrogate under advantages from $R_\phi$ (Eq.~\eqref{eqn:focops_loss}). 
This propagates improvements beyond the exact intervention points but depends on the fidelity of $R_\phi$ and on-policy stability of the RL update for stable and efficient improvement. 
For reference, the proposed SPAR-H combines these two pathways (Sec.~\ref{sec:approach_spar-h}), aiming to retain SPAR-P’s strong local corrections while leveraging SPAR-R’s broader propagation.

\textbf{SPAR-D} applies statewise preferences directly to the intervened states without a reward model, but normalizes logits against a frozen reference policy. This approach adapts from Direct Policy Optimization (DPO, \cite{rafailov2023direct}) for large language models. 
With $\Delta_\theta(s,a)=\log\pi_\theta(a\mid s)-\log\pi_{\text{ref}}(a\mid s)$ the objective becomes
\begin{equation}
\min_\theta\ -\sum \log\sigma\!\Big(\beta\,[\Delta_\theta(s,a^{\text{h}})-\Delta_\theta(s,a^{\text{a}})]\Big).
\end{equation}
We use $\beta=1$ and refresh $\pi_{\text{ref}}$ by snapshot at each epoch. 
Unlike SPAR-H, SPAR-D does not learn a reward model and thus does not perform advantage\mbox{-}based RL.

\textbf{COACH} \cite{macglashan2017interactive}, or COnvergent Actor-Critic by Humans, uses human evaluative feedback as a proxy for the reward advantage on the agent’s proposed action when calculating policy gradients.
\revise{
We include COACH as a representative evaluative RL method that updates the policy directly from scalar human feedback instead of corrective preference, providing a contrast to SPAR-R.
}
With a scalar $f_t\in\{-1,0,1\}$ representing negative, neutral and positive human labels, a one\mbox{-}step update reads
\begin{equation}
\theta \leftarrow \theta + \alpha\, f_t\, \nabla_\theta \log \pi_\theta(a^{\text{a}}_t\mid s_t).
\end{equation}
In our implementation, we omit eligibility traces because feedback is synchronous at each step.
We use binary human labels $f_t\in\{-1,\zeta\}$ to align with the other methods, and include weak positives $\zeta$ ($\zeta=0.1$ in our implementation) on non\mbox{-}intervened steps to gently reinforce safe, uncorrected behaviors ($f_t = \zeta$ if $m_t = 1$).
In contrast to SPAR-H, COACH does not build a reward estimator or convert corrective overrides into preference pairs.

% Reiterate the differences among baseline methods
% IWR and HG-DAgger represent IL methods that supervise the policy on human takeovers but never form preferences or learn a reward. 
% BT and DPO on policy log-probability instantiate direct preference optimization, but update the policy directly rather than via a learned reward. 
% COACH provides a RL baseline that replaces the advantage with a human evaluative signal on the proposed action and likewise omits a reward model. 
% In contrast, CAPER follows the RLHF paradigm: corrective overrides are turned into action-level, same-state preferences to adapt a learned immediate-reward estimator, and the policy is optimized under that reward with a trust-region surrogate. 
% Evaluating across these families isolates the key design axes: corrective vs. evaluative feedback, preference vs. label supervision, and direct-to-policy vs. via-reward pathways.
% Structured comparison of the features of CAPER and the baseline methods for HITL learning are presented in Table \ref{tab:spar_comp}.

These baselines isolate the main design axes we study: corrective vs.\ evaluative feedback, preference vs.\ label supervision, and direct-to-policy vs.\ via-reward pathways. A structured feature comparison appears in Table~\ref{tab:spar_comp}.

\subsection{Learning Framework}
% Image preprocessing 
We represent perception for river following task with a patchified semantic water mask, Fig.~\ref{fig:arch}. 
Each $128 \times 128$ image is segmented by a pretrained model (SAM2 \cite{ravi2024sam2} during deployment) and patchified into a $16 \times 16$ binary grid that encodes river occupancy, aligned with visual pre-processing in \cite{wang2025vision}. 
%This representation is compact, invariant to texture and lighting, and preserves topology along the river corridor. 
% Prior work (Semantic Dynamics Model, \cite{wang2025vision}) showed that such structured, semantics-aligned states support more identifiable dynamics and stronger sim-to-real generalization than latent features learned directly from pixels. 
% The same rationale applies here even though we omit the explicit dynamics head.
% Architecture design
The navigation module consists of a single GRU that consumes the patchified mask and the previous action, producing a latent state. 
Both policy head and reward head are multi-layer perceptrons that share this latent, but output action and estimated immediate reward respectively.
During the forward pass, the policy’s current action is fed into the reward head.
During HITL retraining, the shared GRU is frozen to stabilize the state representation and prevents cross-head interference while the heads adapt to human feedback.
% : the reward loss updates only the reward head and the policy loss updates only the policy head.
% Freezing the recurrent encoder stabilizes the state representation and prevents cross-head interference while the heads adapt to human feedback.
% During forward pass, the current action produced by the policy head is fed back to the reward head. 
% During backpropagation, the reward loss will only affect the reward head, while the policy loss will update both policy head and the shared GRU. 
% This decoupling stabilizes the actor while allowing the reward estimator to adapt to the evolving representation.

\section{EXPERIMENTS}

% \subsection{Simulation}
% Software - Semantic segmentation
In simulation, we use the \textit{medium} level of Safe Riverine Environments (SRE, \cite{wang2024vision}), where ground-truth water segmentation from drone view (facing $15\degree$ downward) is available.
% Drone agent details
The UAV executes high-level, body-frame multi-discrete actions, moving by fixed increments rather than commanding velocities. 
This waypoint-style discretization simplifies control and aligns with river-following objectives without requiring a dynamics controller.
Each timestep selects one value on four branches: vertical translation ($\pm1$ m or 0), yaw ($\pm15 \degree$ or 0), forward/back translation along the nose axis ($\pm1$ m or 0), and lateral translation ($\pm0.5$ m or 0).
The environment constrains motion to a 3-D corridor extruded from the river spline and limits yaw relative to the local spline tangent.
Reward 1 is given when agent visits an un-visited river segment, and episodes end on corridor violations, collisions, or full river traversal. 
% Algorithm implementation
The code is based on the Constrained Actor Dynamics Estimator (CADE, \cite{wang2025vision}) architecture adapted from Safe RL framework Omnisafe \cite{JMLR:v25:23-0681}, but the cost estimator and the dynamics model are inactive to focus on the HITL learning comparison.

% Data collection procedure
To ensure a fair, shared data source and controlled data budget across methods, we use 5 HITL rollouts collected exclusively with a novice policy. 
\revise{
Five HITL rollouts reflect typical field constraints—one battery-limited flight per rollout with a single crew cycle—so operator time and safety oversight remain realistic.
}
At each rollout, the UAV agent will be spawned at a random position with a random pose in the riverine environment.
During rollouts, the overseer followed a conservative intervention policy: intervene only when the proposed action would violate safety (e.g., exit the river spline volume \cite{wang2024synergistic}) or when clear inefficiency was present (e.g., repeated lateral swaying without downstream progress), and otherwise allow the agent’s action to proceed.
Episodes were not length-controlled: each rollout ran until termination, so lengths and intervention rates varied. 
If an episode ended when an unhandled safety violation occurred (e.g., leaving the river spline volume without timely intervention), the terminal transition was excluded from HITL retraining. 
\vspace{-0.1 in}
\begin{figure}[h]
    \centering
    \includegraphics[height=0.18\textheight,width=\linewidth]{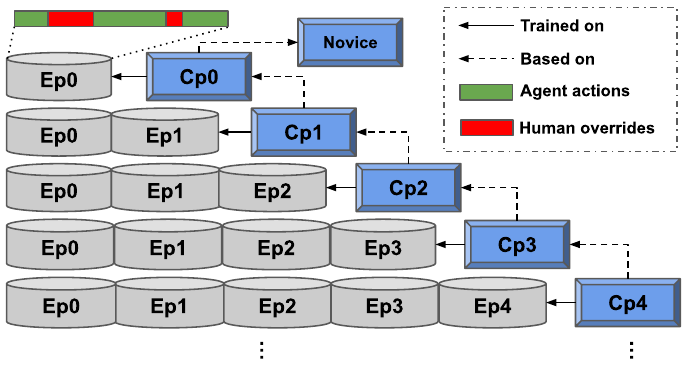}
    \caption{Experiment design for HITL learning. Starting from a novice policy, models are sequentially trained on current existing episodes with human corrective data and saved as checkpoints. Cp stands for checkpoint, and Ep means episode.}
    \label{fig:exp_design}
\end{figure}
\vspace{-0.1 in}

% Experiment design for evaluation
After each episode, we form statewise preference tuples from human interventions and sequentially retrain each HITL method on a cumulative buffer (all episodes seen so far), producing one checkpoint per episode, Fig.~\ref{fig:exp_design}. 
We conduct two evaluations. 
First, after each episode we evaluate the newly retrained checkpoint on the same initial condition as that episode to measure episode-wise improvement from the latest feedback. 
Second, we evaluate the final checkpoint (Cp4) on all five initial conditions to assess overall performance and generalization across starts. 
Episodic reward (task progress before termination) is reported as the primary metric. 
All sources of randomness (environment, policy sampling, and libraries) are seed-controlled for fair comparison.
For hyperparameters, $\alpha = 1$, $\beta = 1$, $\gamma = 0.99$, $\eta = 0.05$, $\lambda = 1.5$, $E = 10$.

% \subsection{Deployment}
% In real-world deployment, the semantic segmentation for the online video stream from the drone uses code \cite{Wang_Segment_Anything_Model} adapted from Segment Anything Model 2 (SAM2, \cite{ravi2024sam2}).
% In real-world deployment, the semantic segmentation for the online video stream from the drone uses code\protect\footnotemark[2] adapted from Segment Anything Model 2 (SAM2, \cite{ravi2024sam2}).

In real-world deployment, the semantic segmentation for the online video stream from the drone uses code\protect\footnotemark[2] adapted from Segment Anything Model 2 (SAM2, \cite{ravi2024sam2}).
% Software - Splashdrone SDK
The SDK\protect\footnotemark[3] of SplashDrone 4 is developed for receiving flight status and video stream, as well as sending control commands via radial frequency. 
% Hardware
% This perception module, Splashdrone 4 SDK, and the SPAR framework are deployed on a Jetson Orin Nano 8Gb Developer Kit.
All the software is deployed on a Jetson Orin Nano 8Gb Developer Kit.
% \footnotetext[2]{The code will be made available upon acceptance.}
\footnotetext[2]{https://github.com/EdisonPricehan/sam2}
\footnotetext[3]{https://github.com/EdisonPricehan/SplashDrone4-SDK}

\section{RESULTS \& ANALYSIS}
\subsection{Simulation}
Across the five rollouts, we collected 846 steps in total with an overall intervention rate of 17.02\%, Table \ref{tab:hitl_demo_stats}.

\begin{table}[h]
\centering
\caption{HITL demonstration statistics.}
\label{tab:hitl_demo_stats}
\begin{tabular}{lrrr}
\hline
\textbf{Episode} & \textbf{Steps} & \textbf{Interventions} & \textbf{Intervention Rate} \\
\hline
0 & 76  & 24 & 31.58\% \\
1 & 456 & 22 & 4.82\%  \\
2 & 158 & 28 & 17.72\% \\
3 & 82  & 31 & 37.80\% \\
4 & 74  & 39 & 52.70\% \\
\hline
\textbf{Overall} & \textbf{846} & \textbf{144} & \textbf{17.02\%} \\
\hline
\end{tabular}
\end{table}

\begin{figure}[h]
    \centering
    \vspace{0.13cm}
    \includegraphics[height=0.2\textheight,width=0.9\linewidth]{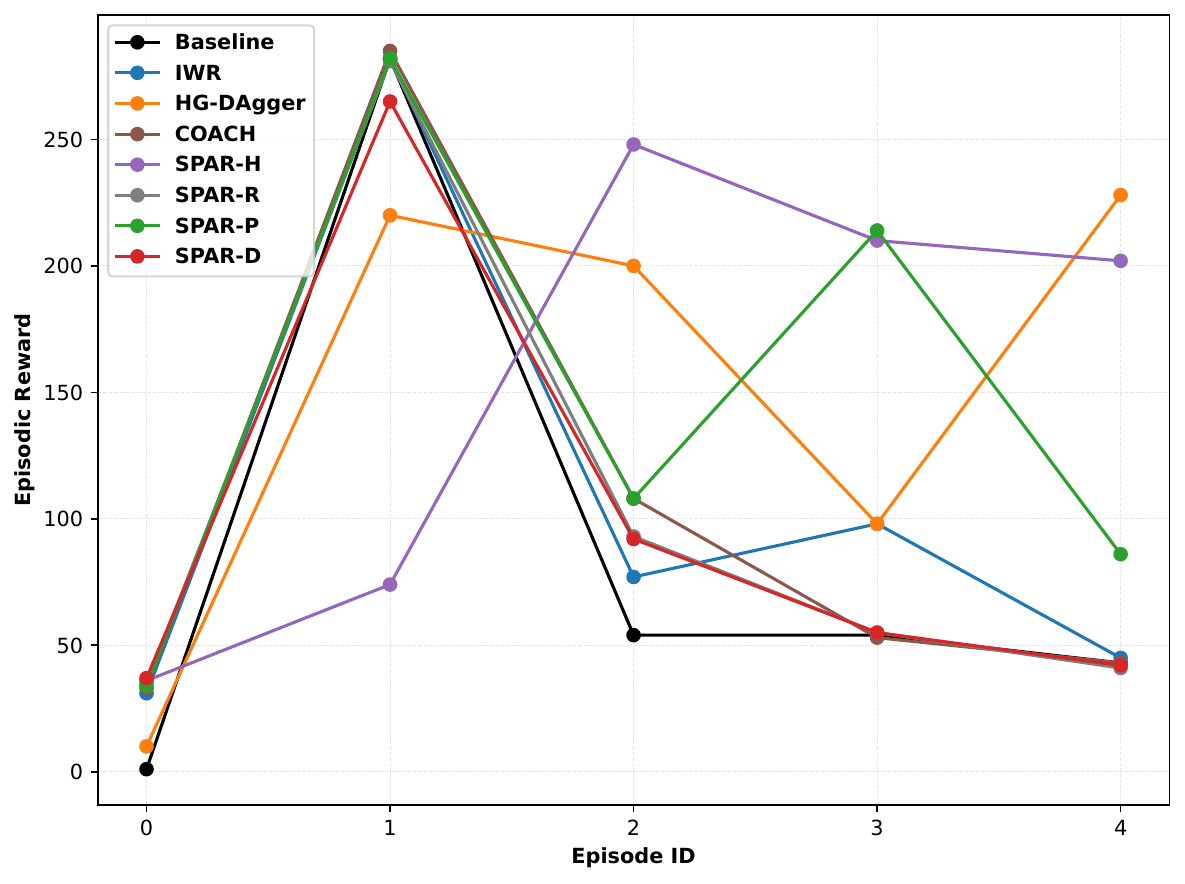}
    \caption{
    % Comparison of episodic rewards across all checkpoints trained via different HITL losses. Each checkpoint is evaluated only at the initial conditions of the episode it was just trained on.
    Episodic rewards per checkpoint. SPAR-H yields the overall largest gains by combining direct and reward-based preference alignment.
    }
    \label{fig:eval_per_ckpt}
\end{figure}

% Analysis of per-ckpt evaluation restuls
Fig.~\ref{fig:eval_per_ckpt} shows the episodic rewards of checkpoints evaluated at the same initial condition of the episode it was just trained on.
This plot presents how each HITL loss converts newly collected human feedback into episode-wise performance gains.
Baseline is the evaluation results of the novice policy.
Relative to the novice baseline, the imitation family (IWR, HG-DAgger) delivers steady gains. 
Notably, IWR outperforms baseline at all five checkpoints, though its margins are smaller than SPAR-H at three points (Cp0, Cp2, and Cp3). 
The direct preference method SPAR-P improves consistently across checkpoints, reflecting efficient in-distribution alignment from BT on policy logits. 
In contrast, the reward-based preference + RL path SPAR-R is stable but mostly flat in gain over the baseline, consistent with conservative FOCOPS updates where the per-state KL gate rejects large steps. 
The hybrid SPAR-H combines SPAR-P’s “surgical” local alignment with SPAR-R’s trust-region propagation and yields the strongest overall performance. 
In practice, tuning the mixing weight $\alpha$ (Eq.~\eqref{eqn:spar-h}) between the two losses should further reduce occasional non-monotonic updates, targeting per-episode improvements that at least match the baseline.

% Comparison of per-episode evaluation results
% \vspace{-0.1 in}

\begin{figure}[h]
    \centering
    \includegraphics[height=0.22\textheight,width=\linewidth]{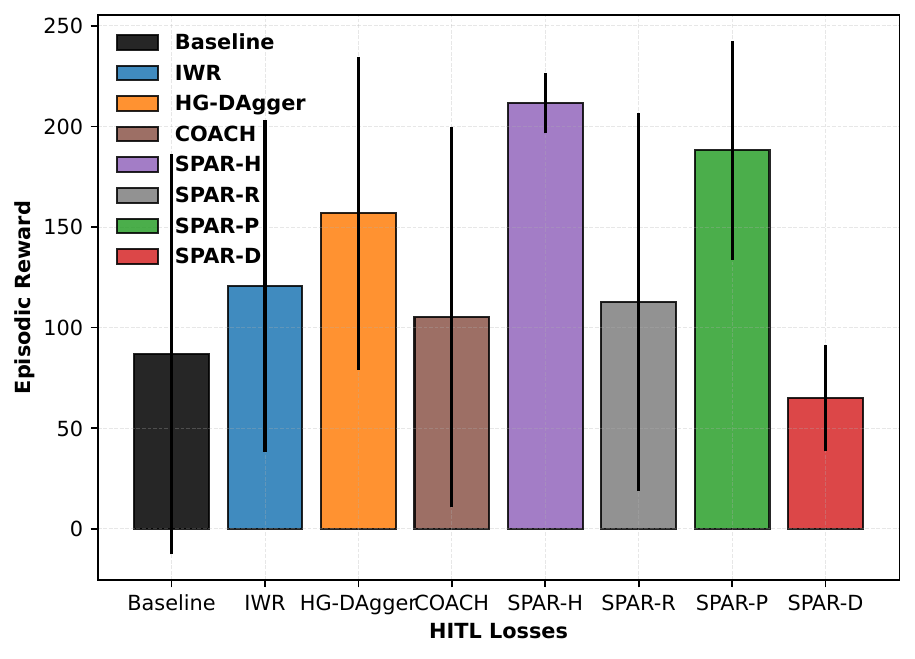}
    \caption{
    % Comparison of averaged episodic rewards of the latest checkpoints (checkpoint 4) trained via different HITL losses. Evaluated for 5 episodes with the same initial conditions as in the collected HITL demo episodes. Vertical line denotes standard deviation.
    Final checkpoint performance. SPAR-H achieves the highest mean reward and lowest variance across initial conditions.
    }
    \label{fig:eval_final_ckpt}
\end{figure}

% Analysis of aggregate ckpt evaluation results
Fig.~\ref{fig:eval_final_ckpt} presents the averaged performance of the latest checkpoints of all loss types evaluated on the same 5 initial conditions.
On the aggregate evaluation, SPAR-H not only achieves the highest mean episodic reward but also the lowest standard deviation, indicating that combining direct statewise alignment with a trust-region RL update yields both strong performance and reliability across initial conditions. 
SPAR-P ranks second, suggesting that using contrastive preferences at intervened states extracts a higher-signal update than either imitation (IWR, HG-DAgger) which lacks negatives, or evaluative RL (COACH) which compresses preference feedback to scalar labels. 
By contrast, DPO (SPAR-D) underperforms here despite being a direct preference method. 
We speculate that, under a small HITL budget and discrete multi-branch actions, the reference-normalized margin can be conservative (yielding small gradients when the reference assigns low mass) and sensitive to the snapshot schedule. 
Overall, the variance patterns mirror these mechanisms: SPAR-H’s trust-region gate stabilizes propagation beyond intervened states, whereas purely direct or purely evaluative paths show larger run-to-run spread.

% Comparison of final checkpoint's eval results on 5 episodes
\vspace{-0.06 in}

\begin{figure}[h]
    \centering
    \includegraphics[height=0.2\textheight,width=\linewidth]{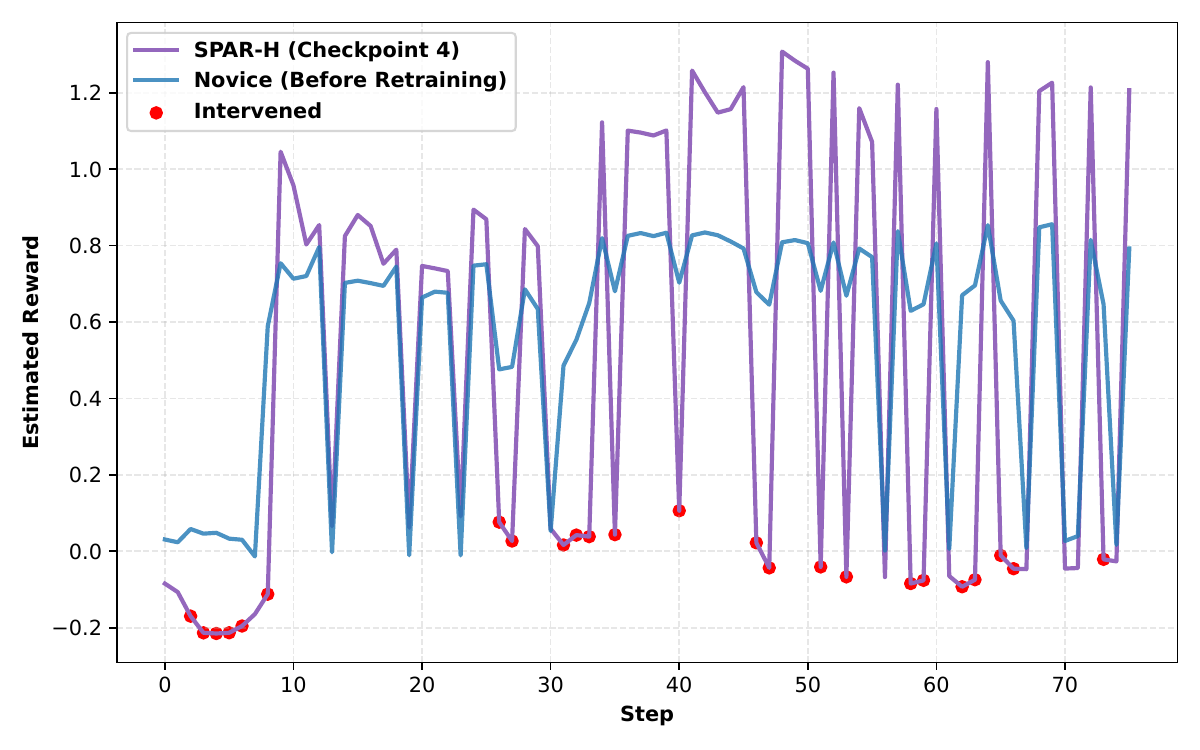}
    \caption{
    % Comparison of the estimated reward between the reward estimator of the novice model and that of the latest checkpoint (checkpoint 4) trained via SPAR-H. Evaluated on the novice-collected trajectory of \textbf{episode 0} where \textbf{agent actions}\protect\footnotemark[1] are adopted for checkpoint rollout.
    Reward estimates of Cp4 (final checkpoint) on Ep0 for $a^\text{a}$. SPAR-H elevates human-approved actions and nearby choices while suppressing rejected ones\protect\footnotemark[4].
    }
    \label{fig:est_rew_comp_agent_act}
\end{figure}

\begin{figure}[h]
    \centering
    \includegraphics[height=0.2\textheight,width=\linewidth]{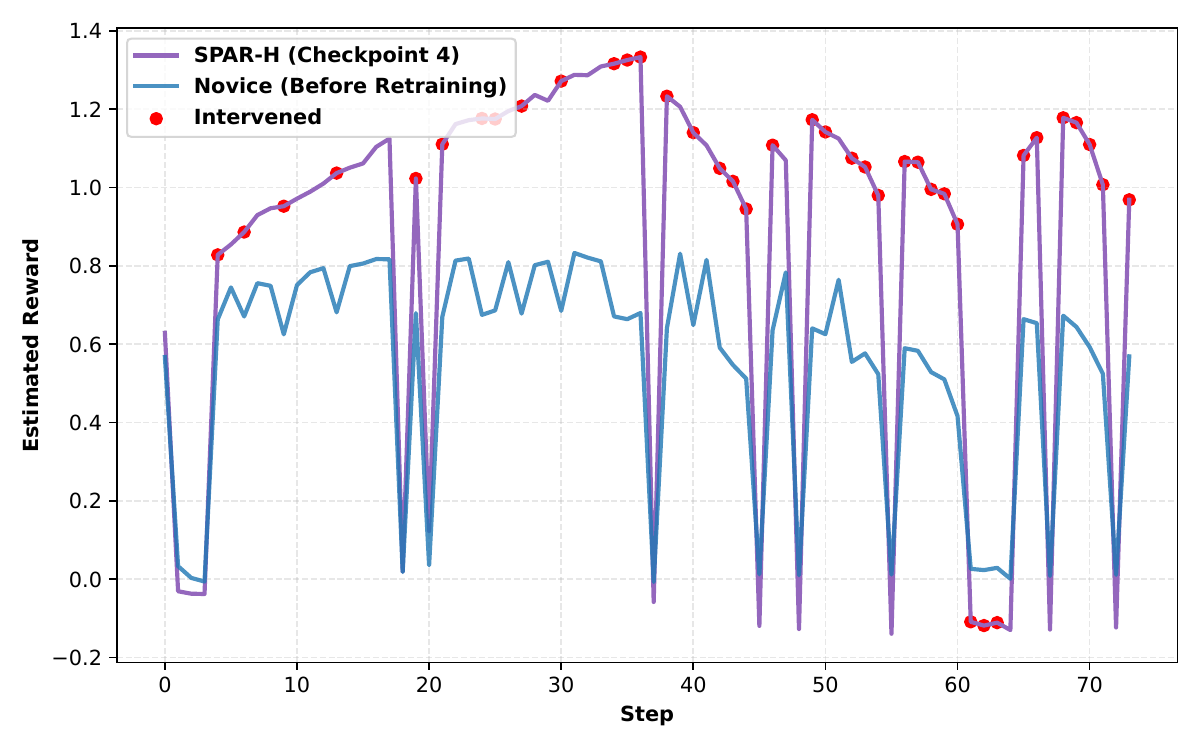}
    \caption{
    % Comparison of the estimated reward between the reward estimator of the novice model and that of the latest checkpoint trained via SPAR-H. Evaluated on the novice-collected trajectory of \textbf{episode 4} where \textbf{executed actions} are adopted for checkpoint rollout.
    Reward estimates of Cp4 on Ep4 for $a^\text{e}$. Human-approved actions form peaks, showing stable alignment over multiple updates.
    }
    \label{fig:est_rew_comp_human_act}
\end{figure}

% Analysis of reward estimator after HITL retraining
Fig.~\ref{fig:est_rew_comp_agent_act} and Fig.~\ref{fig:est_rew_comp_human_act} visualize how incorporating the reward path in SPAR-H reshapes the immediate-reward estimator toward human preferences. 
In episode 0 (agent-proposed actions, Fig.~\ref{fig:est_rew_comp_agent_act}), retraining lifts the estimated reward around human takeovers and depresses competing agent proposals, enlarging the local preference margin compared to the novice model. 
Nearby non-intervened actions are also elevated, indicating short-range propagation of the corrective signal beyond the exact intervention state. 
By episode 4 (executed actions), these effects persist after several updates: peaks cluster around human-approved choices, troughs deepen around rejected behaviors.
Together, the figures show that retaining and updating the reward model online lets preferences shape not only the intervened points but also their neighborhood, which is valuable when the policy is later optimized under this learned reward.
\vspace{-0.2cm}
% Comparison of estimated rewards before and after HITL retraining
\footnotetext[4]{Note: $s_t$ is computed from the executed history, so we plot $R_\phi(s_t,a^{\mathrm{a}}_t)$ without assuming the ensuing transition.}
\vspace{-0.1 in}

\subsection{Real-World Deployment}
% Real world experiment
The UAV was deployed on the Wabash River (Indiana, USA) from Davis Ferry Bridge to Highway~65, covering approximately 1.7\,km. 
We conducted five battery-limited sorties, producing five trajectories and 377 total steps.
The waypoint controller was initialized with the same novice policy used in simulation and updated sequentially across batteries, carrying forward the latest weights to the next trajectory.
To extend range, waypoint increments were set to 10\,m forward, 5\,m lateral, and 5\,m vertical. 
A human overseer intervened conservatively, overriding actions only when safety was at risk or the vehicle was evidently inefficient.
For faster adaptation, SPAR-H retrained the model every 10 steps using the most recent interventions. 
Episodes with no interventions did not trigger a retrain.

\begin{figure*}[h]
    \centering
    \vspace{0.15cm}
    \includegraphics[height=0.18\textheight,width=0.99\linewidth]{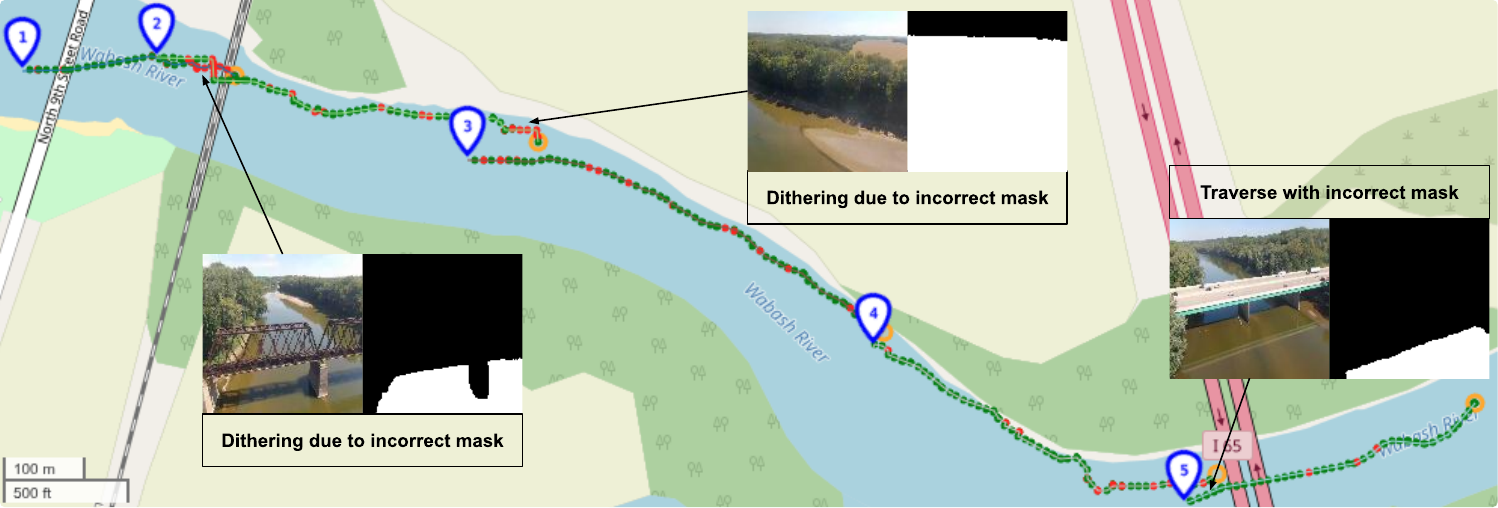}
    \includegraphics[height=0.13\textheight,width=\linewidth]{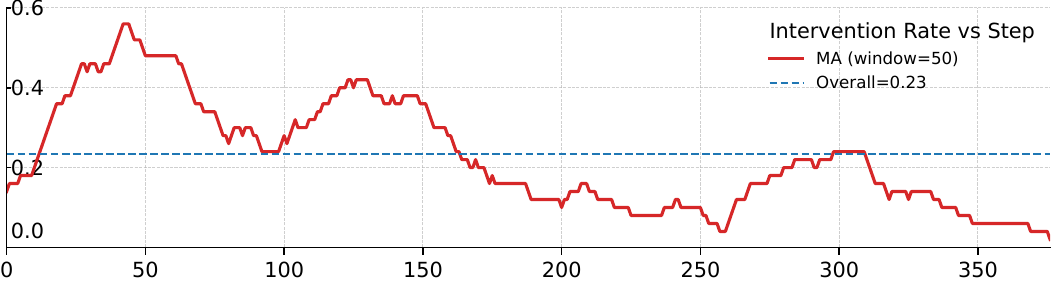}
    \caption{Five HITL trajectories during deployment with SPAR-H. Green dots: executed agent-proposed actions. Red dots: human overrides. Orange dots: trajectory ends.
    \revise{Bottom: moving averaged intervention rate per 50 steps.}
    Interventions taper across sorties as the policy adapts online.}
    \label{fig:wabash_trajectories}
\end{figure*}

% Present results and analyse
Figure \ref{fig:wabash_trajectories} shows five HITL trajectories with online adaptation via SPAR-H. 
The densest interventions occur late in Traj-1 and Traj-2, where water-mask errors beyond the bridge induced dithering: repeated rotations stalled progress.
Later runs require only sparse, mostly efficiency-oriented corrections.
By Traj-5, the agent clears the highway autonomously without intervention despite the same segmentation issue seen in Traj-1. 
\revise{
The moving average intervention rate drops below 0.23 per 50 steps in the second half of river navigation.
}
Overall, intervention severity and frequency decrease across sorties, even under imperfect perception, indicating rapid preference alignment and safer, more efficient autonomy.

% Figure of trajectories and intervention rates

\section{CONCLUSIONS}
We investigated HITL learning for vision-driven UAV river following and introduced SPAR-H, a statewise hybrid that couples direct preference optimization on policy logits with a reward-based path trained from the same preferences and updated via RL.
Under the shared HITL rollouts with limited intervention budget, SPAR-H achieves the highest final episodic reward and the lowest variance across episodes.  
The learned immediate-reward estimator shifts toward human-preferred actions and elevated nearby non-intervened choices, showing evidence of local alignment and stable propagation. 
Comparisons against imitation learning, direct preference variants, and evaluative RL, together with a feasibility demonstration in real-world deployment, indicate that dual statewise preferences offer a practical and data-efficient route to online adaptation.
\revise{
While demonstrated for river following, SPAR-H generalizes to other safety-critical partially observable domains such as urban flight corridors or ground vehicle navigation.
}
Future work could study the sensitivity to the mixing weight between paths and to trust-region thresholds, active preference querying, and uncertainty-aware preference models.
% Limitations include sensitivity to the mixing weight between paths and to trust-region thresholds, and reliance on the quality of human feedback. Future work will study adaptive weighting, active preference querying, uncertainty-aware preference models, and larger-scale field trials with tighter safety constraints.


%%%%%%%%%%%%%%%%%%%%%%%%%%%%%%%%%%%%%%%%%%%%%%%%%%%%%%%%%%%%%%%%%%%%%%%%%%%%%%%%

%%%%%%%%%%%%%%%%%%%%%%%%%%%%%%%%%%%%%%%%%%%%%%%%%%%%%%%%%%%%%%%%%%%%%%%%%%%%%%%%

%%%%%%%%%%%%%%%%%%%%%%%%%%%%%%%%%%%%%%%%%%%%%%%%%%%%%%%%%%%%%%%%%%%%%%%%%%%%%%%%
%\section*{APPENDIX}


% \section*{ACKNOWLEDGMENT}

%%%%%%%%%%%%%%%%%%%%%%%%%%%%%%%%%%%%%%%%%%%%%%%%%%%%%%%%%%%%%%%%%%%%%%%%%%%%%%%%

% \newpage
\bibliographystyle{IEEEtran}
\bibliography{reference.bib}

\end{document}